\documentclass[conference]{IEEEtran}
\IEEEoverridecommandlockouts

\makeatletter
\let\old@ps@headings\ps@headings
\let\old@ps@IEEEtitlepagestyle\ps@IEEEtitlepagestyle
\def\confheader#1{%
	\def\ps@headings{%
		\old@ps@headings%
		\def\@oddhead{\strut\hfill#1\hfill\strut}%
		\def\@evenhead{\strut\hfill#1\hfill\strut}%
	}%
	\def\ps@IEEEtitlepagestyle{%
		\old@ps@IEEEtitlepagestyle%
		\def\@oddhead{\strut\hfill#1\hfill\strut}%
		\def\@evenhead{\strut\hfill#1\hfill\strut}%
	}%
	\ps@headings%
}
\makeatother

\confheader{%
	2020 28\textsuperscript{th} Iranian  Conference on Electrical Engineering (ICEE)
}
\usepackage[pscoord]{eso-pic}
\newcommand{\placetextbox}[3]{
	\setbox0=\hbox{#3}
	\AddToShipoutPictureFG*{ \put(\LenToUnit{#1\paperwidth},\LenToUnit{#2\paperheight}){\vtop{{\null}\makebox[0pt][c]{#3}}}
	}
}
\placetextbox{.23}{0.055}{\small{978-1-7281-7296-5/20/\$31.00~\copyright 2020 IEEE}}

\usepackage{cite}
\usepackage{amsmath,amssymb,amsfonts}
\usepackage{algorithmic}
\usepackage{graphicx}
\usepackage{textcomp}
\usepackage{xcolor}
\usepackage{color, colortbl}
\usepackage{multirow}
\usepackage[left=1.57cm,right=1.57cm,top=1.9cm,bottom=2.54cm]{geometry}
\def\BibTeX{{\rm B\kern-.05em{\sc i\kern-.025em b}\kern-.08em
		T\kern-.1667em\lower.7ex\hbox{E}\kern-.125emX}}
\usepackage{footmisc}
	
\begin{document}
	
	\title{An Autonomous Intrusion Detection System Using an Ensemble of Advanced Learners}
	
	\author{\IEEEauthorblockN{Amir Andalib }
		\IEEEauthorblockA{\textit{School of Electrical Engineering} \\
			\textit{Iran University of Science and Technology}\\
			Tehran, Iran \\
			amir\_andalib@elec.iust.ac.ir}
		\and
		\IEEEauthorblockN{Vahid Tabataba Vakili}
		\IEEEauthorblockA{\textit{School of Electrical Engineering  } \\
			\textit{Iran University of Science and Technology}\\
			Tehran, Iran \\
			vakily@iust.ac.ir}
			}
	
	\maketitle
	
	\begin{abstract}
		An intrusion detection system (IDS) is a vital security component of modern computer networks. With the increasing amount of sensitive services that use computer network-based infrastructures, IDSs need to be more intelligent and autonomous.  Aside from autonomy, another important feature for an IDS is its ability to detect zero-day attacks. To address these issues, in this paper, we propose an IDS which reduces the amount of manual interaction and needed expert knowledge and is able to yield acceptable performance under zero-day attacks. Our approach is to use three learning techniques in parallel: gated recurrent unit (GRU), convolutional neural network as deep techniques and random forest as an ensemble technique. These systems are trained in parallel and the results are combined under two logics: majority vote and "OR" logic. We use the NSL-KDD dataset to verify the proficiency of our proposed system. Simulation results show that the system has the potential to operate with a very low technician interaction under the zero-day attacks. We achieved $87.28 \%$ accuracy on the NSL-KDD's "KDDTest+" dataset and $76.61 \%$ accuracy on the challenging "KDDTest-21" with lower training time and lower needed computational resources.
	\end{abstract}


	\section{Introduction}
	\label{introduction}

	Computer networks' settled and pervasive role in all aspects of our daily lives must ring a bell for its serious security vulnerabilities. Since all of our information, from identification to where we go, what we like, our medical history, consumption pattern, etc. go through these networks, any kind of vulnerability can lead to an irrecoverable disaster.
	
	To deal with security challenges in computer networks, many methods have been introduced: cryptography and firewalls are of such efforts. One of the robust and reliable systems is intrusion detection system (IDS). From an operational point of view, there are mainely three types of IDSs: 1. Misuse-based, 2. Anamoly-based 3. Hybrid (uses both misuse and anomaly techniques)\cite{survey1}.

	Despite many efforts to entrust security challenges to a system without constant human interaction, we are still in the beginning steps of building such systems. To make anomaly-based IDSs intelligent, artificial intelligence and its promising subfield, machine learning, are being widely used.  
	Developing paradigms and methods of machine learning are at the intense focus of computer science and other related research fields such as mathematics. Other fields of science and researches benefit from its various techniques. The challenge toward applying it as a powerful tool to a special problem is to choose the best method and to set its parameters. Evidence suggests the trial-and-error procedure is a powerful way to find the best method, especially in deep learning methods; since it is currently unknown what makes a deep learning structure to work finely suited 
	\cite{emergingids}.
	
	After choosing the proper technique, there are additional challenges: First, at any level of the learning procedure, having deep insight into the subject can be very crucial. Feature selecting, preprocessing and making data ready for training is another challenge and somehow the most important step in the procedure. Finally, one tricky challenge is the system's ability to generalization i.e. the trained system strength against inputs which are not very similar to the training data. The later challenge can be interpreted as detecting zero-day attacks in the context of network intrusion detection.
	
	In this work, we are trying to address the first and last challenges by proposing a network IDS, based on using several learning methods. First, network packets for a single connection are stored and analyzed with tools such as: "Snort" and "Bro IDS" where a set of features is extracted. Then, to classify the type of the connection as normal or attack, we use a system trained  with three different methods: a recurrent neural network (RNN) with gated recurrent unit (GRU) base, a convolutional neural network (CNN) and random forest (RF). Using different types of classifiers and combing their votes with proper logic shows that we can have an IDS which is able to detect various zero-day attacks without manually adjusting the learning procedure.   When the system encounters a misclassified attack, it learns to cope with it in the future.  We have evaluated our system on the KDD-NSL dataset which is a well-known and most commonly used dataset in the field. The rest of this paper is organized as follows. In Section \ref{background}, a brief background about in-use techniques is provided. In Section \ref{relatedworks}, we review some of the related researches. Section \ref{systemarchi} provides a detailed description of our proposed system. In Section \ref{experimental}, we discuss the experimental results. Finally, the paper concludes in Section \ref{conclusion}.
	
	\section{Background}
\label {background}
In this section, we state three needed preliminaries: 1. A brief description of the used classifiers. 2. Preprocessing phase. 3. In-use dataset.


\subsection {Random Forest}
By using ensemble learning method, random forest (RF) is introduced to overcome some issues of decision trees (DTs) such as overfitting and lack of criterion for feature selecting. The training dataset is randomly divided into several subsets and each subset trains different DTs. Any of these trained trees claims their votes and then majority vote method is applied \cite {rf}.
A theoretical background given by L.Breimann in \cite{rf} indicates that RFs always converge and overfitting is not a problem anymore.

\subsection{RNN with GRU unit}
In contrast to a classical neural network which only has uni-direction connections, RNN  takes advantage of the recurrent connection between layers.
As a deep neural network, RNN  has two major issues known as "vanishing gradient problem" and "exploding gradient". As the number of layers grow, the gradients of loss function  become too small or too large.
This makes the training procedure difficult or even impossible. One of the most effective ways to tackle the aforementioned problems is to use the GRU unit. GRU is introduced as a simplified version of Long Short-Term Memory (LSTM - another memory-based unit). The memory cell is the core of a GRU unit, which allows it to maintain state over time\cite{geron}. 




\subsection {Convolutional Neural Networks}
Convolutional neural networks (CNNs) try to mimic the brain's visual cortex functionality by defining a variety of filters which results in extracting various information from images. Although the visual perception is their main task, evidence implies that they are effective in other learning-related tasks as well.
LeNet-5, AlexNet, GoogleNet, and ResNet are some popular CNN architectures. A CNN architecture is a special combination of layers, namely convolution, pooling and fully connected layers.

\subsection{Preprocessing}
To prepare data for training, two common steps are numericalization and normalization of the features. Numericalization is assigning numbers to nominal features. In addition, to avoid unbalanced effects of the features  in classification, they are all normalized to the interval  $[0,1]$  by eq. (\ref{normalization_eq}), in which $f_i$ is the feature value:

\begin {equation}
\label{normalization_eq}
f_{i,nrlz.} = \frac{f_i-f_{min}}{f_{max} - f_{min}}.
\end {equation}

\subsection {Dataset description}
For  training and evaluating an anomaly-based IDS, a reliable and genuine dataset is needed. To have a valuable dataset, background flow integrity and attack variety are considered to be the most important properties.  
NSL-KDD is a purified version of KDD99 dataset for the KDD (international Knowledge Discovery and Data mining tools competition) cup, presented by Tavallee and et al. in \cite{nsl}. 
This dataset was introduced in 2004 and since then many things in the Internet and computer network field have been changed. However, due to its wide usage,  it is still regarded as the best reference dataset for comparing several IDS systems performance. It has three predefined datasets, KDDTrain+ for training, KDDTest+ for testing and a challenging KDDTest-21 set. They have designed 21 machine learning systems for evaluating their proposed dataset and the KDDTest-21 is a subset of KDDTest+ in which, any records classified correctly by all of the 21 machines, are removed \cite{nsl}. 

	\section{Related works}
 	\label{relatedworks}
 	Here, we mention some of the papers that implement or optimize NIDS. 
 	By using three search techniques (genetic algorithm (GA), ant colony optimization (ACO), and particle swarm optimization (PSO)) B.A. Tama et al in \cite{tse} propose a  feature selection method. They apply this search method to a two-stage ensemble classifier which uses rotation forest and bagging methods. The evaluation of this method on two datasets is done and the results are: $85.79 \%$ accuracy on the KDDTest+ set and $91.27 \%$ accuracy on the UNSW-NB15 set. 
 	In \cite {dcgan}, J. Yang et al propose an IDS using the deep convolutional generative adversarial network. Instead of analyzing network packets with a network monitor software, their proposed system  directly extracts features from raw data and generate a new dataset. They also apply a modified long short-term memory (LSTM) which is simple recurrent unit (SRU) allowing the system to learn the important features of intrusions. They achieve $99.73 \%$ accuracy on the KDD99 dataset and  $99.62 \%$ on manually divided NSL-KDD dataset (The predefined subsets of NSL-KDD dataset are not used for training and testing.). 
 	S.M. Kansongo et al. in \cite{ffdnn} use a filter-based algorithm for a wireless IDS to select features and a feed-forward deep neural network. Their neural network has 3 hidden layers with 30 nodes. They achieve 99.69\% accuracy on NSL-KDD dataset by employing $0.05$ learning rate.
 	In \cite{rnn} C.Yin et al. propose a RNN-based IDS. They use a classical RNN without a memory unit and by examining the range of 20 to 240 hidden nodes for their RNN, at 80 nodes and the learning rate $0.1$, they achieve 83.28 \% and 68.55\% accuracy on KDDTest+ and KDDTest-21, respectively.
 	By using a feature selection algorithm called modified binary grey wolf optimisation (MBGWO), Q.M. Albuzi et al. in \cite{tse21}, address some challenges of the feature selection phase. They use support vector machine as the classifier and evaluate their work on NSL-KDD dataset (accuracy of 81.58\% on KDDTest+).
	\section {Proposed System Architecture}
 	\label{systemarchi}
 	Detecting zero-day and new types of anomalies are the most sophisticated work that can be done by IDSs. Designing such a system and keeping it up to date and fast need simultaneous maintenance. To address the challenge of lowering permanent interaction, we propose an IDS  by combining the strength of three different types of machine learning techniques and designing an update procedure. In the following subsections, we depict the IDS operation manner in the network and in the training course.
 	
 	\subsection{System Operation Phase} 
\begin{figure}
 		\centering
 		\includegraphics[scale=0.43]{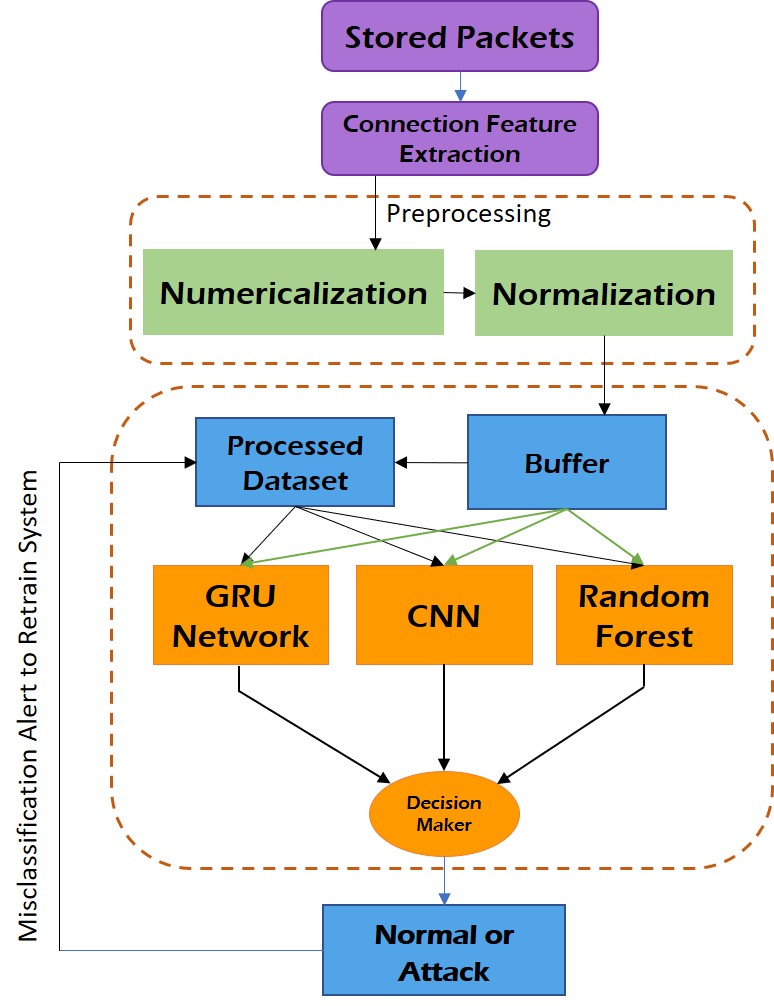}
 		\caption{Proposed IDS Operation Scheme }
 		\label {proposed_sys}
 	\end{figure}

As shown in Figure \ref{proposed_sys}, first, packets of a connection are stored, then a machine that is capable of executing a packet analyzer (here we use "Zeek", i.e. an open-source software which can be run on Raspberry PI), extracts features based on predefined rules. Then,  three trained machines do the classification and claim their votes to a decider. The decider combines votes based on the assigned logic. 
 	
 	We use three different types of learning techniques to establish the IDS classifier unit. There are some parameters and some assumptions which motivates us to choose these three techniques. 
 	We  discuss the rationale behind setting the number of subsystems to  three later in this section.

\begin{figure}
 		\centering
 		\includegraphics[scale=0.45]{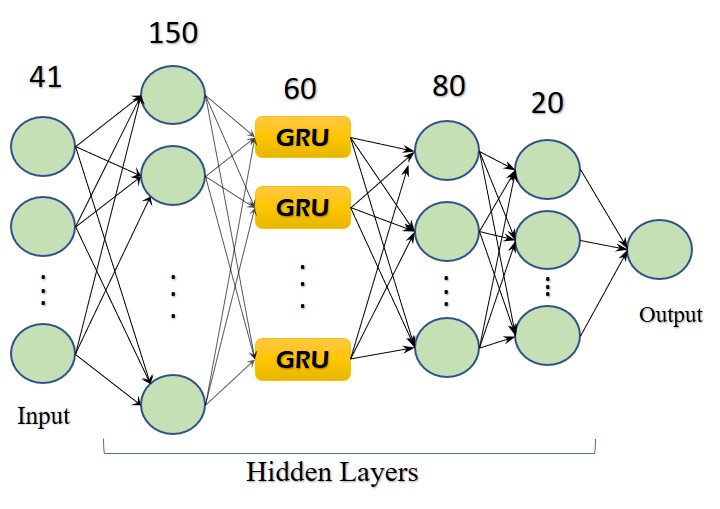}
 		\caption{Proposed GRU Network }
 		\label {my_gru}
 	\end{figure}

 	One of our in-use techniques is the GRU-based network. As mentioned earlier, RNN networks have the ability to remember previous entries. This means that we can use RNN as a time analysis tool. In many modern and sophisticated attacks, malicious codes can be injected in distributed patterns, for example by using botnets or embedding the codes into many legal packets. Hence the aforementioned ability of RNN turns out to be very effective in the processing of packets sequences. Figure \ref{my_gru} shows our proposed GRU network architecture. 
 	To tackle vanishing gradient problem, we use "Leaky ReLU" and exponential linear unit (ELU) activation functions given by the equations \cite{geron}:  
 	
 	\begin {equation}
 	\label{leakyrelu}
 	g(x) = \begin{cases} 
 		x ,  & x \geq 0\\
 		\alpha x ,& x < 0
 	\end{cases}
 	\end {equation}
 	and
 	\begin {equation}
 	\label{elu}
 	g(x) = \begin{cases} 
 		x ,  & x \geq 0\\
 		\alpha (\exp^x-1) ,& x < 0,
 	\end{cases}
 	\end {equation} respectively. The number of layers and their nodes are adjusted based on an exhaustive search (an example is given in Section \ref {experimental}).

\begin{figure}
 		\centering
 		\includegraphics[scale=0.4]{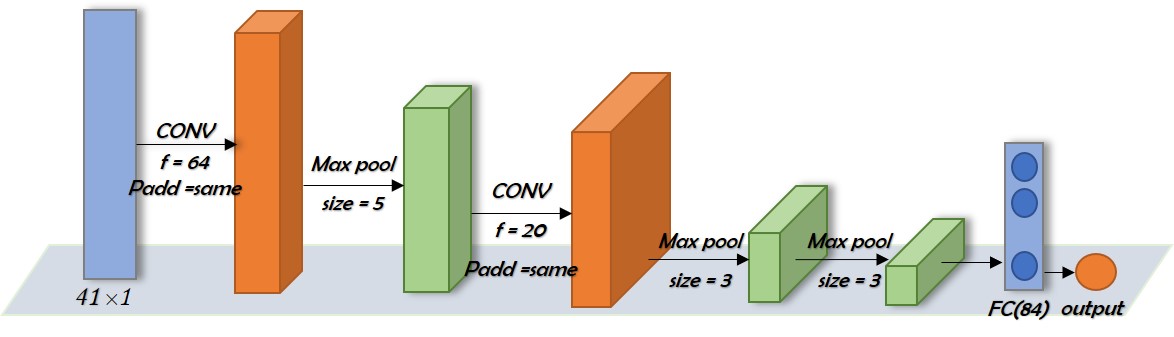}
 		\caption{In-Use LeNet-5 Structure}
 		\label {my_cnn}
 	\end{figure}

 	The second learning machine of our proposed system is a CNN network. 
 	CNN networks are used for their competence to extract features. On the other hand, IDSs operate in different types of networks that have different dynamics and threats. Thus, to increase the autonomy of the IDS, automatic feature extraction is needed. Therefore, a CNN is embedded in the system. Here, we use a modified version of one of the earliest and proficient networks (LeNet-5). The modification is deduced from simulation results.  Figure \ref{my_cnn} shows the structure of our applied version of LeNet-5.
 	
 	Our last learner is a RF that is claimed to be robust against overfitting. Overfitting reduces the generalization attribute of a classifier, thus IDS may lose the ability of zero-day attack detection. Other reasons for using this technique are its speed and nonrandom training procedure.
 	
 	Our concluded key characteristics for implementing an IDS with the aforementioned attributes are having  time analyzer, memory and feature extractor within the machines.
 	After many efforts by implementing several learning techniques, even unsupervised methods, we find the solution in using GRU and CNN. The randomness of these two deep methods obliges us to use a more robust classifier (RF).

 	\subsection{Training Phase}
 	In order to train system, first, the data collected from a dataset is processed and gets ready to be inserted into classifiers, then every machine is trained in the parallel mode. Now the machine is ready to operate. During its operation, it checks its prediction accuracy. When there is an uncaptured attack, the feedback procedure informs the training system. It adds the misclassified connection record to the training dataset and performs the training again. 

 	%
	
	\section {Experimental Results}
	\label{experimental}
	The hardware and software setups used in our experiment are as follows: 
	\begin {itemize}
	\item {Intel Core i7 @ 3.5Ghz with 16 GB of RAM. (No GPU-Based Implementation)}
	\item {Tensorflow v.2.0.0b1 and scikit-learn v0.21.3 libraries on CentOS v7.5 operation System}
\end{itemize}

\subsection {Performance Evaluation}
The effectiveness of classification methods is measured by metrics such as accuracy (Acc), detection rate (DR) and false positive rate (FPR). These metrics are constructed on the following four items: 1. True positive(TP)  2. True negative (TN), 3. False positive (FP) and 4. False negative (FN). The evaluation metrics are defined as follows:
\begin{itemize}
\item {Accuaracy: The percentage of total records that is  classified correctly (\ref{eq_acc}):
	\begin {equation}
	\label{eq_acc}
	Acc = \frac {TN + TP}{TN + FN + FP + TP}.
	\end {equation}
}

\item{Detection Rate : DR is the percentage of correctly identified attack records (\ref{eq_dr}):
	\begin{equation}
		\label{eq_dr}
		DR = \frac {TP}{TP+FN}.
\end{equation}}

\item{False Positive Rate : FPR is the ratio of incorrectly attack alarms to all incorrect identifications
	(\ref{eq_fpr}):
	\begin{equation}
		\label{eq_fpr}
		FPR = \frac {FP}{FP + TN}.
\end{equation} }
\end{itemize}

\begin{table*}  [h!]
\tiny
\caption{Random Forest Results for different estimator numbers}
\label{results_rf}
\begin{center}
	\begin{tabular}{|l| c|c|c|c|c|c|c|c|c|c|c|}
		\hline
		No. of Estimators   &10	     &20	  &	30 	   &	40   &50     &60     &70	   &80	    &90	     & 100 & 200\\
		\hline
		Train Acc Mean 		&99.974& 99.976& 99.981& 99.979& 99.982&$\mathbf{ 99.983}$& 99.983&99.981& 99.982& 99.983& 99.982 \\
		Train Acc Std.		&0.0033& 0.0028& 0.0017& 0.0012& 0.0007&$\mathbf {0.0008}$& 0.0009&	0.0005& 0.0005& 0.0004& 0.0003 \\
		\hline
		Valid Acc Mean		&99.873& 99.833& 99.841& 99.833& 99.817& \cellcolor{yellow!25}$\mathbf{99.849}$& 99.865&99.873& 99.849& 99.849& 99.849\\
		Valid Acc Std. 		&0.0197& 0.022 & 0.0158& 0.0162& 0.014 & \cellcolor{yellow!25}$\mathbf{0.0134}$& 0.0158&0.0147& 0.0114& 0.0117& 0.0102\\
		\hline
		Test Acc Mean		&80.424& 79.812& 79.617& 80.251& 79.741& $\mathbf{79.954}$& 79.785&79.861& 80.034& 80.078& 80.171\\
		Test Acc Std.		&0.543& 0.493& 0.637& 0.344& 0.395& $\mathbf{0.33}$ & 0.469& 0.494& 0.405& 0.41 & 0.308 \\
		\hline
		Test-21 Acc Mean	&62.996& 61.595& 61.451& 62.523& 61.603& $\mathbf{62.008}$& 61.747&	61.772& 62.068& 62.211& 62.338 \\	
		Test-21 Acc Std. 	&1.23&0.942& 1.194& 0.648& 0.723& $\mathbf{0.67}$ & 0.895& 0.928& 0.784&0.794& 0.584  \\
		\hline 
		Learning Dur.(s) 	& 3.181&  5.693&  8.482& 10.944& 13.457&$\mathbf{16.322}$& 19.064& 21.28 & 24.176& 26.854& 52.3 \\
		\hline

	\end{tabular}	
\end{center}
\end{table*}

\begin{table}[h]
\tiny
\caption {Summary of the Proposed System Configuration}
\label{brief_config}
\begin{center}
	\begin{tabular}{|l|l|c|}
		\hline
		\multirow{10}{*}{CNN / GRU} & Hidden Act. Func. & Leaky ReLU \& ELU \\ \cline{2-3}
		& Output Act. Func. & Sigmoid\\\cline{2-3}
		& Batch Size 	& 1024\\\cline{2-3}
		& No. of Epochs & 100\\\cline{2-3}
		& Learning Rate & 0.006\\\cline{2-3}
		& Cost Func. 	& Binary CrossEntropy\\ \cline{2-3}
		& Optimizer 	& Adam\\ \cline{2-3}
		& Bias			& Yes\\ \cline{2-3}
		& Regularization & No\\ \cline{2-3}
		& Class Weighting & No \\ \cline{2-3}
		
		\hline
		\multirow{2}{*}{RF}  	   & No. of Estimators & 60\\\cline{2-3}
		& Max Features			& Auto (equal to no. of features)\\\cline{2-3}
		& Criterion & gini\\
		\hline
	\end{tabular}	 					    
\end{center}
\end{table}
\subsection {Results on Binary Classification}
We consider a  variety of configurations for each method of learning to find out the best combination. For instance, the number of nodes and GRU units, optimizer, learning rate, activation function, class weighting, regularization, and initialization are set to different values to find the best configuration. Some definite results are: 1. "Adam" optimizer always outperforms the stochastic gradient descent (SGD), "RMSProp", "adagrad" and "adadelta".  2.  Although regularization is claimed to be helpful to control high bias fitting, here we do not observe any considerable changes in the accuracy results of the KDDTest-21 dataset.  
We implement the same structure for 30 times and observe the mean and standard deviation of the accuracy results. We choose a combination that has a better mean with low standard deviation for validation set with rational learning time. Table \ref{results_rf} shows the results for RF with different numbers of estimators. The standard way to validate the performance of a learner is to argue on the validation set. It is obvious that with high numbers of estimators the standard deviation of the validation set accuracy tends to be lower with almost constant accuracy values, but the learning time grows. We set numbers of estimators to $60$ which has the lowest standard deviation  with proper learning time (highlighted cell in Table \ref{results_rf}). The same procedure\footnote{https://github.com/catcry2007/nsl4conf \label{github}} is applied to GRU network and CNN  with different numbers of units and filters, respectively  and we choose $50$ GRU units which has the average accuracy $99.72 \%$ with   $0.04\%$ standard deviation. The average learning time of GRU-based is $91.73$ $(sec)$. Although we set the epochs to 100 but  an early stop callback mechanism is embedded to stop learning procedure when the loss of the validation set  $(val\_loss)$  does  not descend  anymore with the patience of four epochs (or when it grows up). The latter mechanism also helps improve the generalization attitude of the classifier.  Table \ref{brief_config} shows our applied configuration briefly.  All of the codes, detailed results for different neural network configurations and the best obtained weights corresponding to their layers are accessible in our "github" repository\footref{github}.

\begin{figure}[t]
\centering
\includegraphics[scale=0.40]{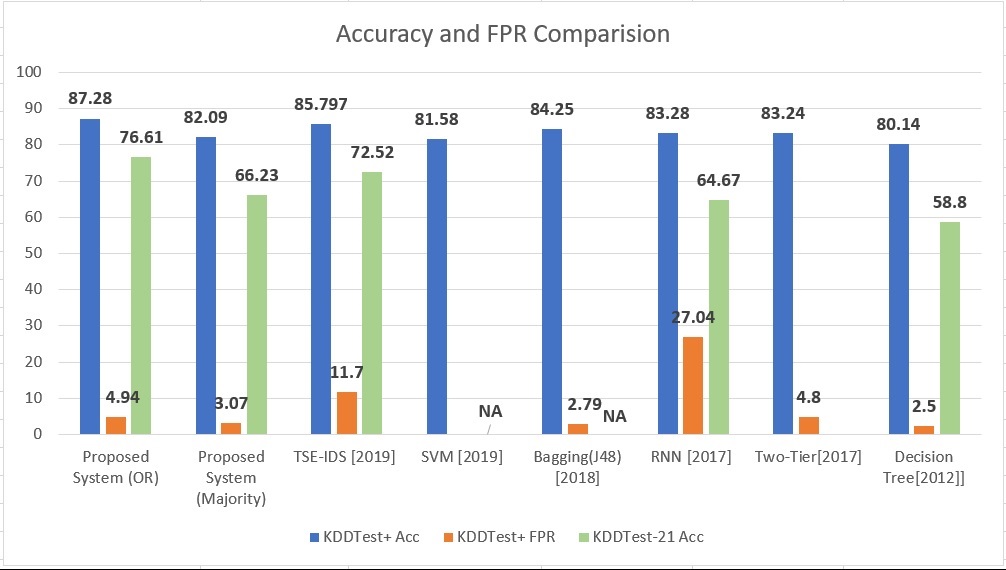}
\caption{Comparison of TSE-IDS \cite{tse}, SVM \cite{tse21}, Bagging (J48)\cite{tse25}, RNN\cite{rnn}, Two-Tier \cite{tse49}. }
\label {predefined_res}
\end{figure}

As mentioned earlier, the NSL-KDD composes of 3 predefined sets that some of the attack types are not included in the train set; so the system can be evaluated for zero-day attacks. Most of the papers including \cite {dcgan,ffdnn ,gru} combine the predefined sets and apply a regular learning procedure, i.e. dividing the dataset to the training, validation and test sets . Obviously, in the latter case, the performance would be higher. We evaluate our system with the predefined sets and the comparison with other methods are shown in Figure \ref {predefined_res}. Figure \ref{predefined_res} shows that our proposed system improves the accuracy of the tricky  KDD-Test-21 set by more than $4\%$ comparing to the latest best achievement \cite{tse} and by $2 \%$ in KDD-Test set. It is worth mentioning that we reduced the training time by about 617 seconds (from about $830$ $(sec)$ in \cite{tse} to $212.5$ $(sec)$).
Table \ref{detailed_results} shows the detailed results of sets' accuracy and training time for each subsystem of our proposed system.	
Table \ref{conf_matrix_test} and \ref{conf_matrix_test_21} show the confusion matrix of the our system for the KDDTest+ and KDDTest-21 sets, respectively.   	

\begin{table}[h!]
\scriptsize
\caption {Subsystems training time and accuracy for the validation, KDDTest+ and KDDTest-21 sets}
\label{detailed_results}
\begin{center}
	\begin{tabular}{|l|c|c|c|c|}
		\hline
		sub-sys  & Validation & KDDTest+ & KDDTest-21 & Training time\\\hline
		CNN				&  99.63 			&82.92		      &	68.30			  &104.45(sec)  \\\hline
		GRU				& 99.76				&83.19			  &		68.52	      & 91.73(sec)\\\hline
		RF				& 99.79				&80.14		      &		62.34	      & 16.32(sec)\\
		\hline
		
	\end{tabular}	 					    
\end{center}
\end{table}					

\begin{table} [h] 
\scriptsize
\caption {Confusion Matrix of the "KDDTest+" set}
\label{conf_matrix_test}
\begin{center}
	\begin{tabular}{l|c|c|}
		\multicolumn{1}{c}{}		& 	\multicolumn{1}{c}{Predicted Normal}	& \multicolumn{1}{c}{Predicted Attack} \\ \cline{2-3}
		Normal	&	9230	&	480\\ \cline{2-3}
		Attack	& 	2387	&	10446 \\ \cline{2-3}
		
	\end{tabular}	 					    
\end{center}
\end{table}
\begin{table}  [h]
\scriptsize
\caption {Confusion Matrix of the "KDDTest-21" set}
\label{conf_matrix_test_21}
\begin{center}
	\begin{tabular}{l|c|c|}
		\multicolumn{1}{c}{}		& 	\multicolumn{1}{c}{Predicted Normal}	& \multicolumn{1}{c}{Predicted Attack} \\ \cline{2-3}
		Normal	&	1769	&	383\\ \cline{2-3}
		Attack	& 	2388	&	7310 \\ \cline{2-3}
		
	\end{tabular}	 					    
\end{center}
\end{table}	

%
%
%
	\section{Conclusion and Future Direction}

	\label{conclusion}
	
	In this paper, we proposed an IDS that is able to operate with the minimum interaction in training and updating procedure and performed acceptably well under zero-day attack detecting. Our proposed IDS is able to update the dataset and learn to deal with new misclassified records. We examined this IDS with the NSL-KDD dataset and the results showed improvement in terms of both accuracy and training time duration compared to state-of-the-art methods. There are still many challenges to be defined, such as:  1. Running the system under real-world networks, 2. Adding the ability to detect the attack type.

	\bibliographystyle{IEEEtr}
	\bibliography{refz}

\end{document}